# An Enhanced Computational Feature Selection Method for Medical Synonym Identification via Bilingualism and Multi-Corpus Training

Kai Lei, Shangchun Si, Desi Wen and Ying Shen *
Institute of Big Data Technologies
Shenzhen Key Lab for Cloud Computing Technology & Applications
School of Electronics and Computer Engineering(SECE)
Peking University, SHENZHEN 518055 P.R.CHINA
Email: leik@sz.pku.edu.cn, danieljay@126.com, wendesi@foxmail.com
Corresponding Author: *shenying@pkusz.edu.cn*

*Abstract*—Medical synonym identification has been an important part of medical natural language processing (NLP). However in the field of Chinese medical synonym identification, there are problems like low precision and low recall rate. To solve the problem, in this paper, we propose a method for identifying Chinese medical synonyms. We first selected 13 features including Chinese and English features. Then we studied the synonym identification results of each feature alone and different combinations of the features. Through the comparison among identification results, we present an optimal combination of features for Chinese medical synonym identification. Experiments show that our selected features have achieved 97.37% precision rate, 96.00% recall rate and 97.33% F1 score.

## I. INTRODUCTION

Nowadays medical literature and data is growing rapidly. How to efficiently obtain valuable information from these massive data becomes an issue of growing concern but challenging. In the medical domain, one same concept may have different ways of expression. For instance, the Chinese word "肝癌"(liver cancer) has synonyms like "肝肿瘤","肝细胞癌","肝腺瘤", and English synonyms like "liver cancer" , "liver tumor", "liver carcinoma", "hepatic carcinoma" etc.

Medical synonym identification is important in the field of medical NLP [1]. Synonym extraction is important in constructing a high quality medical NLP system. It helps a lot in improving the precision and coverage rate of results in query expansion, text summarization, question answering, entity alignment and paraphrase detection. However manual construction of medical synonym dictionary is always expensive, which may lead to low knowledgebase coverage, especially in the medical domain. Moreover, the natural language content is growing at an extremely high speed in the medical domain, making people hard to understand it, and hard to update it in the knowledgebase timely.

Compared with the English medical terms, there's less study and lack of sophisticated algorithms and reliable data on Chinese medical synonym identification. The identification of Chinese synonyms is thereby with low precision rate and low recall rate.

In this paper, we propose a method for identifying Chinese medical synonyms. Assisted by English translation information, this method makes full use of the semantic information captured by word embedding model in large-scale bilingual corpus and takes advantage of the unique structure and pronunciation characteristics of Chinese characters to identify the synonyms in the Chinese medical domain.

The contributions of this paper can be outlined as follows:

- Compared with other work in Chinese medical synonym identification, the 13 selected features are more comprehensive. Through experiment we evaluate the performance of each feature alone, and furthermore, their performance in different combinations by enumeration computation. At last, among all the 13 features, we present the best feature combination for Chinese medical synonym identification.

- Experiment results reveal the following conclusion: Semantic features obtained through word embedding and search engine information do well in the identification. Phonetic (pinyin) information can to some extent eliminate the difference of transliteration. Ideographic (radical) information makes up for the lack of morphological information of other features, and brings about quite good results.

*Resrach supported by School of Electronics and Computer Engineering (SECE), Peking University Shenzhen Graduate School.

## II. RELATED WORK

There are many approaches to extract and identify synonyms, including lexical patterns, Wikipedia and search engine-based methods.

Lexical pattern-based methods are first used to extract word semantic relationships by Hearst [2]. He applied patterns like "A such as B" to detect hypernym-hyponym relationships. Based on this, Wang et al. [3] proposed an automatic pattern construction approach by using some seed synonyms (antonyms), which were extracted from WordNet using heuristic rules. But these methods require good linguistics understanding to put up the patterns and rules.

Wikipedia-based methods are often used because Wikipedia contains abundant (semi-)structured knowledge, which is useful for synonym identification. Weale et al. [4] proposed an approach to identify synonyms using the graph structure of Wiktionary. Based on a direct measure of information flow in the graph and a comparison of vertices close to a given vertex, the semantic relation is measured.

Search engine-based methods are also used with the popularity of search engines and the accumulated search log, which are useful corpus for mining information. By using similarity functions named "click similarity", webpage URLs and queries that have clicked these URLs are identified.

However the Wikipedia-based and search engine-based methods neglect information contained in the plain text itself like morphological features.

Specifically, in medical domain, professional medical ontology is often used to assist synonym identification and similarity measuring. Mainly there are two approaches of calculating similarity: edge-based and node-based [5].

In edge-based method, the number of paths between two terms in the graph is calculated. Similarity is measured by the shortest path or the average length of all paths. Besides, semantic similarity can be calculated by the path length from the lowest common ancestor to the root node. Node-based method relies on the attributes of the terms. In these methods, semantic similarity can be measured by the information entropy of their common ancestor. Compared to other edge-based methods, entropy-based method is less sensitive to changeable semantic distance.

Liang [6] proposes one way to calculate the similarity between diseases by measuring the gene similarity based on gene ontologies. Wang et al [13] presents a novel approach for medical synonym extraction, aiming to integrate the medical domain knowledge with the term embedding for further applications.

But the limitation of these methods is obvious: they all need professional medical ontology for support.

## III. METHOD

In machine learning, support vector machines (SVM) are supervised learning models with associated learning algorithms for data classification and regression analysis. Given a train set, each sample is marked for belonging to one of two categories. The SVM training algorithm constructs a model that assigns new samples into one category or the other. An SVM model represents the examples as points in space that are mapped so that the samples of the separate categories are divided by an explicit gap which is as wide as possible. New samples are then mapped into that same space and predicted to belong to a category based on which side of the gap they fall on.

In this paper, SVM is used to identify whether a pair of words is synonyms by means of the selected features, and the precision rate, the recall rate and the F1 score are calculated respectively. In total, 13 Chinese and English features are selected, including 3 word vector features, 6 word-level features, 2 Chinese-specific features, and 2 semantic-level features.

For the better indication of the experiment results, these features are named from Feature 1 to Feature 13 respectively below.

### A. Chinese Synonym Determination Assisted by English Translation Information

English translation is adopted to help determining whether a term pair is a pair of synonyms. Bilingual synonym extraction algorithm is common means [7] [8]. As far as we know, there are few studies exploring the combination of different resources for synonym extraction. However, many studies investigate synonym extraction from only one resource.

Each Chinese medical term in the dataset is translated into a set of English terms, recorded as $Enlist_{1,2}$. All of the translated results are kept.

### B. Word Vector Calculation

With the idea of deep learning and developments in distributional semantics, neural network-based approaches get more and more popular. These approaches use a three-layer neural network consisting of an input layer, a hidden layer and an output layer. The neural network itself models the language model and obtains a representation of words in the vector space [9] [10].

To train the model only requires a large amount of unlabelled text data. This data is used to create a semantic space. And terms are represented in this semantic space as vectors that are called word embeddings (word vectors).

These approaches leverage the context information of a word, and greatly enrich semantic information. The geometric properties of this space prove to be semantically meaningful. Words that are close in the semantic space tend to be semantically similar [9] [10].

Due to the rich semantic information the word vectors contain, neural network-based approach is adopted in this paper. In this way the process of textual content can be simplified into vector operations in the semantic space.

*Cosine Similarity (Feature 1)*

The similarity in the word vector space can be used to represent the semantic similarity of text. Cosine similarity measures similarity between two vectors by computing the cosine of the angle between them, namely the inner product. Cosine similarity of the two multidimensional word vector A, B is defined as follows:

$$CosSim(A,B) = \frac{A \cdot B}{\|A\|\|B\|} = \frac{\sum_{i=1}^{n} A_i B_i}{\sqrt{\sum_{i=1}^{n} A_i^2} \sqrt{\sum_{i=1}^{n} B_i^2}} \quad (1)$$

It can be seen that the cosine of 0° is 1, and it is not greater than 1 for any other angle. It is thus a judgment of orientation but not scale: two vectors with the same orientation have a cosine similarity of 1; two vectors at 90° have a similarity of 0; and two vectors diametrically opposed have a similarity of -1, independent of their scale. The bigger the value, the more similar the two words are.

Similar to [1], in our work, given two translated terms set EnList1 and EnList2, we estimate their similarity by using the average vector of EnList:

$$AvgVec = \frac{1}{n} \sum_{i=1}^{n} term_i, term_i \in EnList \quad (2)$$

Cosine similarity of two sets (Feature 2), represented by two average vectors can be calculated using formula (1).

*Euclidean distance (Feature 3)*

In mathematics, the Euclidean distance is the "normal" distance between two points in Euclidean space. Apparently, smaller Euclidean distance means higher similarity.

For the obtained word vector, Euclidean distance can also be used to calculate the similarity, which is defined as follows：

$$EuDist(A,B) = \sqrt{\sum_{i=1}^{n}(A_i - B_i)^2} \quad (3)$$

## C. Word-level Features

*Edit Distance*

In computer science, edit distance [11] is a way of measuring how different two strings (e.g., words) are. It counts the minimum number of operations (insertion, deletion and substitution) required to transform one string into the other. Generally, the less the edit distance is, the more similar the two strings are. Edit distance is an effective method to calculate similarity in medical text processing [12].

To normalize the measure, relative edit distance (Feature 4) is defined below, where maxLength (A, B) is the max length of the two strings A, B:

$$EditDist(A,B) = \frac{editDistance(A,B)}{maxLength(A,B)} \quad (4)$$

As for translated results, relative edit distance between two sets Enlist1, 2 (Feature 5) is:

$$EnEditDist = Max\{EditDist(A_i, B_j)\}$$

$$A_i \in EnList_1, B_j \in EnList_2 \quad (5)$$

*English Morphological Feature*

In linguistics, morphology is the study of the internal structure of words, and the rules by which words are formed, and their relationship to other words in the same language. English speakers can recognize the relations between words from their tacit knowledge of English's rules of word formation.

Several features are adopted including:

Duplicate word (Feature 6): Returns 1 if there are duplicate words in two translated sets; otherwise, 0.

Subsequence (Feature 7): If one term is the subsequence of another term, returns 1; otherwise, 0.

First character (Feature 8): If all the first characters in each word from a and b match each other, returns 1; otherwise, 0[13]. For example, "liver cancer" and "liver carcinoma" sharing the same first characters "lc", then returns 1.

Abbreviation (Feature 9): If all the upper case characters from a and b match each other, returns 1; otherwise, 0. For example, m4 =1 for "USA" and "United States of America".

## D. Chinese-Specific Information

*Pinyin Edit Distance (Feature 10)*

Pinyin[1] is unique to the Chinese language. Similar to English phonetic symbol, pinyin is used as phonetic representation of Chinese characters. Chinese characters can be similar to each other because of the same or similar sound. So pinyin edit distance can be used to calculate the Chinese word similarity [14]. There are many transliterated ones in the Chinese medical domain. Pinyin can eliminate the difference of transliteration. For instance, the pinyin of "埃博拉病毒" 和"埃播拉病毒" (both meaning Ebola virus) is the same. So pinyin edit distance is used as a feature.

Pinyin information is obtained through the Xinhua Chinese Dictionary, and changes of the four tones are ignored. Similar to 3.1, the relative edit distance between the two pinyin sequences is calculated.

*Number of radicals (Feature 11)*

As hieroglyphics, Chinese can be similar when the characters look alike. Chinese radicals[2] have their specific meaning, and through which Chinese achieve its ideographic meaning. So the words with same radicals generally have more similar meaning [15]. For Chinese word pairs, the more common radicals they share, the more similar they are. Meanwhile in the medical domain, radicals appearing at a high frequency such as "月" (part of the body), "艹" (some kind of bacteria), "疒"(some disease or virus), have a high degree of discrimination. It makes up for the lack of ideographic meaning of other features.

Number of common radicals (CR) is defined below, which is divided by max length of A, B for normalization.

---

[1] https://en.wikipedia.org/wiki/Pinyin
[2] http://en.wikipedia.org/wiki/Radical_ (Chinese_characters)

$$CR(A,B) = \frac{commonRadicals(A,B)}{maxLength(A,B)} \quad (6)$$

The radical information is obtained from the Xinhua Chinese Dictionary.

*E. Semantic-level features*

The Normalized Google Distance [16] is a measure of semantic similarity. For a given set of keywords, it is derived from the hit number returned by Google. Keywords with the same or similar meanings in a natural language sense tend to be "close" in the Normalized Google Distance unit, while words with different meanings are likely to be far apart. Nowadays as there are huge amount of resources on the Internet are medical related. Search engine information can be an effective measure of semantic similarity in medical domain.

For two search terms x and y, the Normalized Google Distance (NGD) (Feature 12) is defined as follows:

$$NGD(x,y) = \frac{max\{log f(x), log f(y)\} - log f(x,y)}{log M - min\{log f(x), log f(y)\}} \quad (7)$$

M is the total number of web pages searched by Google (here, we set logM = 10); f(x) and f(y) are the hits number of search terms x and y respectively; f(x, y) is the number of web pages where x and y occur at the same time.

If the two words always appear at the same time, then NGD(x,y) = 0, which means x and y are viewed as alike as possible; if NGD(x,y) ⩾ 1,then x and y are very different; if the two search terms x and y never appear together on the same page, but appear separately, then NGD(x, y) is infinite.

Baidu is the world's largest Chinese search engine. Similar to Normalized Google Distance, Normalized Baidu Distance (Feature 13) is defined in this study with the same formula above.

## IV. RESULT

*A. Datasets and Corpus*

In this paper, in order to build train and test dataset, we construct a Chinese medical thesaurus in a semiautomatic way. In this progress, synonym data are mainly obtained from the medical related pages of online encyclopedias (e.g. Wikipedia, Hudong Encyclopedia and Baidupedia) via web crawlers. Meanwhile, the synonym data also come from: A plus medical encyclopedia [1] and Xunyiwenyao [2], (most professional medical websites in China) and authoritative online medical dictionaries. Terms are restricted to disease names and symptoms.

Finally, 2882 pairs of synonyms were obtained as positive samples. As for negative samples, 2882 terms (same amount of positive samples) are randomly selected from the modern Chinese dictionary together with medical terms to form term as negative pairs. Among all the 5764 word pairs, two thirds are randomly selected as train set, and the rest one third are used as test set. We keep the ratio of positive to negative as 1: 1 in both train and test sets.

Acquisition of word vectors requires large-scale corpus for training the model. Chinese corpus is derived from the text of the online encyclopedia mentioned above, as well as the medical knowledge base text data; Inspired by the paper [17], English corpus is derived from DailyMed, WikiDisease, WebMD, and MayoClinic.

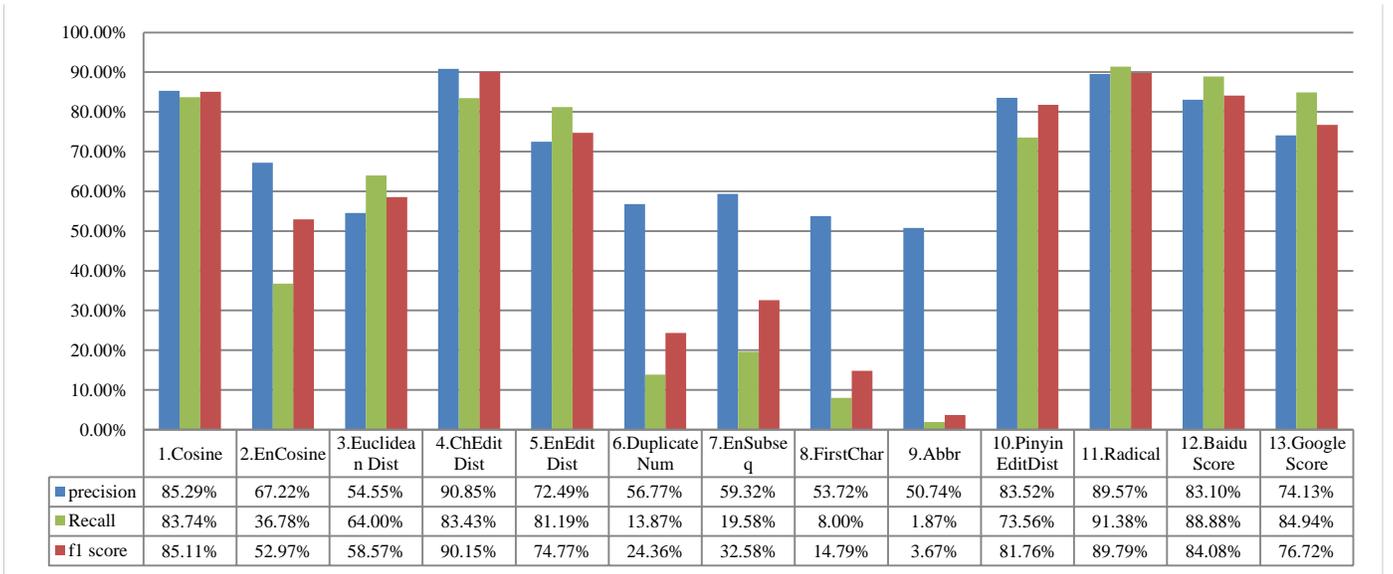

| | 1.Cosine | 2.EnCosine | 3.Euclidean Dist | 4.ChEdit Dist | 5.EnEdit Dist | 6.Duplicate Num | 7.EnSubseq | 8.FirstChar | 9.Abbr | 10.Pinyin EditDist | 11.Radical | 12.Baidu Score | 13.Google Score |
|---|---|---|---|---|---|---|---|---|---|---|---|---|---|
| precision | 85.29% | 67.22% | 54.55% | 90.85% | 72.49% | 56.77% | 59.32% | 53.72% | 50.74% | 83.52% | 89.57% | 83.10% | 74.13% |
| Recall | 83.74% | 36.78% | 64.00% | 83.43% | 81.19% | 13.87% | 19.58% | 8.00% | 1.87% | 73.56% | 91.38% | 88.88% | 84.94% |
| f1 score | 85.11% | 52.97% | 58.57% | 90.15% | 74.77% | 24.36% | 32.58% | 14.79% | 3.67% | 81.76% | 89.79% | 84.08% | 76.72% |

Figure 1. Results when using each feature exclusively

---

[1] http://www.a-hospital.com/
[2] http://www.xywy.com/

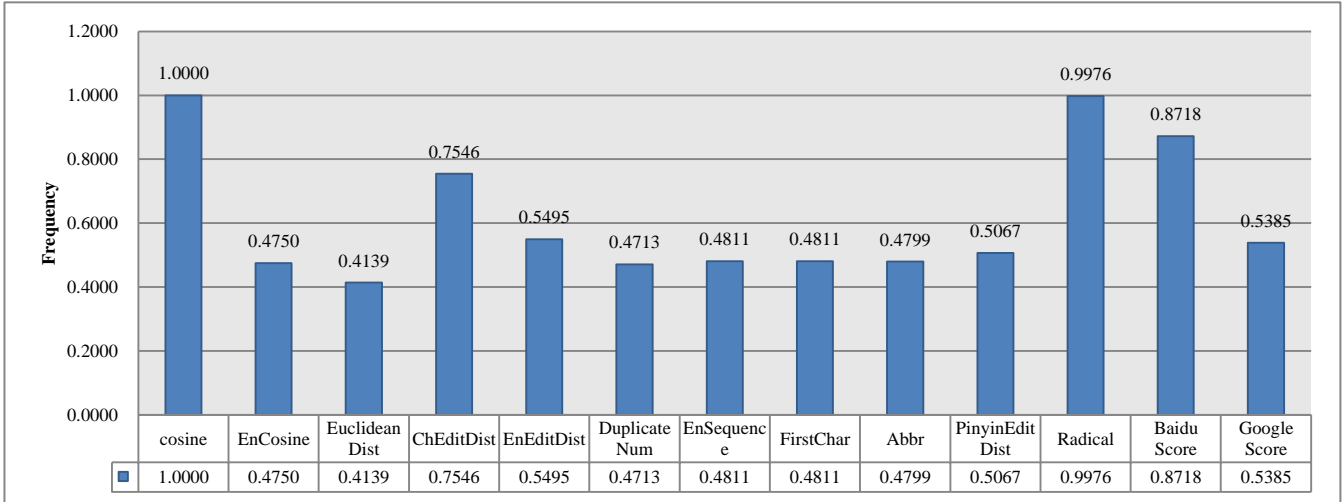

Figure 2. Frequency of Each Feature in top 10% Combinations Sorted by F1 Score

A Plus Medical Online Dictionary [1], which provides translations of several professional medical dictionaries, is used for translation of the medical terms. As for those non-medical terms, Youdao online dictionary [2] is used for translation.

### B. Performance of Each Feature Alone

In this part, we study the performance of each feature alone. The results for each feature are shown in Fig.1.

In terms of precision, Chinese edit distance and radical do the best, while radical and Baidu score get the highest recall rate. As for F1 score, Chinese edit distance, radical cosine and Baidu score are the best ones. Besides, our results reflect the fact that when searching Chinese medical terms, Baidu does better than Google.

Among the thirteen features, cosine similarity, search engine score, and unique Chinese features bring about the best results. In the medical domain, Chinese characteristics (edit distance, radical information and pinyin edit distance) have a good classification effect; however, the number of abbreviations, first characters, and the duplicate words are less effective for medical synonym identification

### C. Performance of Each feature in Combinations

Then we study the performance of each feature in the feature combinations. The best features are Chinese cosine distance based on word vector, Chinese-specific radical information, edit distance both in English and Chinese, Normalized Google and Baidu Distance, and pinyin edit distance (Feature 1,4,5,10,11,12,13) .

In our preliminary experiments, the best combination of results varies in different sets of data, but generally speaking, the combinations of features that yield better results in single feature test above tend to achieve good effects. So we do the following experiment to check if it is true.

In order to observe the contribution of each feature, and identify which combinations of features are more effective, we performed $2^{13}$ experiments with or without one or more specific features (For the 13 features, to use or not to use each feature, there are $2^{13}$ cases in total). In each case, precision rate, recall rate and F1 score are calculated respectively.

In the $2^{13}$ combinations, we select the top 10% (sorted by F1 score) to observe the frequency of each feature. The result is shown in Fig. 2. Features with higher frequency means they perform better in the combinations.

From the figure we can see the best features are ones mentioned above.

### D. Overall Performance

As for the overall performance of feature combinations, Table.1 shows the best ten results in $2^{13}$ feature combinations, sorted by F1 score. Corresponding relationship between numbers and features is shown in Fig. 1. The best result is the combination of cosine similarity, English cosine similarity, English edit distance, pinyin edit distance, common radicals, Baidu and Google score (feature No. 1,2,4,7,9,11,12,13), where the precision is 97.37%, the recall rate is 96.00%, the F1 score is 97.33%.

Table 1． Results of top10combinations of features (sorted by F1 score)

| Feature combinations | precision | Recall | f1 score |
|---|---|---|---|
| 1,2,4,7,9,11,12,13 | 97.48% | 96.31% | 97.44% |
| 1,4,5,9,11,12,13 | 97.48% | 96.21% | 97.44% |
| 1,2,4,6,11,12,13 | 97.48% | 96.21% | 97.44% |
| 1,4,5,8,11,12,13 | 97.48% | 96.21% | 97.44% |
| 1,2,3,4,5,6,11,12 | 97.48% | 96.10% | 97.44% |
| 1,3,4,5,6,11,12 | 97.48% | 96.10% | 97.44% |
| 1,2,3,4,5,6,8,11,12 | 97.48% | 96.10% | 97.44% |
| 1,2,4,5,11,12 | 97.48% | 96.00% | 97.43% |
| 1,4,5,11,12 | 97.48% | 96.00% | 97.43% |
| 1,2,4,11,12,13 | 97.42% | 96.21% | 97.39% |

[1] http://www.mcd8.com/
[2] http://fanyi.youdao.com

To verify the stability and effectiveness of the feature combination, we run the experiment with the feature combination for ten times. Each time the train and test data set are randomly selected. The result shows that the feature combination brings about stable and good precision, recall and F1 score. The experimental result is omitted here.

We take related work in recent two years for comparison. Wang et al. [13] from IBM Watson Research Lab aims to integrate the term embedding with the medical domain knowledge for medical synonyms. The best result of theirs is 70.97% in terms of F1score (precision and recall rate are not mentioned in the paper), which is outperformed by our result by a large margin. The paper [18] from Tsinghua University and Carnegie Mellon University studied the normalization of Chinese medical terms. Our result surpasses theirs, whose best precision is 0.907(recall and F1 score are not mentioned).

Besides, the results of Table 1 verify the accuracy of Fig. 1 from the other side. At the same time, the results reveal the possibility of using English translation information to assist identity Chinese synonyms. The bilingual edit distance is useful in the Chinese medical synonyms identification. Besides, the English morphological features reflect the improvement of bilingual recognition of synonyms.

## V. Conclusion and Future Work

In this paper, we propose a method for identifying Chinese medical synonyms. Compared with other work, the 13 selected features are more comprehensive. Assisted by English translation information, this method makes full use of the semantic information captured by word embedding model in large-scale bilingual corpus and takes advantage of the unique structure and pronunciation characteristics of Chinese characters to identify the synonyms in the Chinese medical domain. Search engine scores also do well.

We also evaluate the performance of each feature alone, and their performance in different combinations by enumeration computation. At last, among all the 13 features, we present the best feature combination for Chinese medical synonym identification.

In conclusion, the feature combination for identifying Chinese medical synonyms are: Chinese cosine distance based on word vector, Chinese radical information, edit distance both in English and Chinese, normalized Google and Baidu distance, and pinyin edit distance.

In the future work, experiment will be focused on more specific fields like symptoms, drugs and disease names to further enhance the accuracy of synonym identification. We also aim extract Chinese characters that are distinguishable in the medical field, such as "阳性"(positive), "阴性"(negative), "病"(disease) and so on to see if they can achieve better classification results.

At the same time, more features like SNOMED and other medical terminology lexicons will be added. We also plan to enlarge the data set and apply the same methods, to check the performance of our method.